\documentclass[11pt]{article}

\usepackage[preprint]{acl}

\usepackage{times}
\usepackage{latexsym}

\usepackage[T1]{fontenc}

\usepackage[utf8]{inputenc}

\usepackage{microtype}

\usepackage{inconsolata}

\usepackage{graphicx}
\usepackage{xcolor}
\usepackage{amsmath}
\usepackage{caption}    
\usepackage{subcaption}
\usepackage{booktabs}
\usepackage{cleveref}
\usepackage{algorithm}
\usepackage{algpseudocode}
\usepackage{cleveref}

\usepackage{todonotes}
\usepackage[normalem]{ulem}

\makeatletter
\newcommand*\iftodonotes{\if@todonotes@disabled\expandafter\@secondoftwo\else\expandafter\@firstoftwo\fi}  
\makeatother

\newcommand{\layer}[1]{{\textcolor{purple}{#1}}}       
\newcommand{\doc}[1]{{\textcolor{blue}{#1}}}           
\newcommand{\tok}[1]{{\textcolor{orange}{#1}}}            
\newcommand{\lang}[1]{{\textcolor{teal}{#1}}}          

\usepackage{framed}
\usepackage{color}
\definecolor{shadecolor}{RGB}{230,240,255}

\title{Leveraging Routing Dynamics in Mixture-of-Experts Models \\for Efficient Language Adaptation}

\author{
Aditi Khandelwal ~~ Marius Mosbach ~~ Verna Dankers ~~ Siva Reddy ~~ Golnoosh Farnadi \\
\\
Mila – Quebec AI Institute \& McGill University \\
\small{\texttt{\{aditi.khandelwal, marius.mosbach, verna.dankers, siva.reddy, gfarnadi\}@mila.quebec}}}
\makeatletter
\AddToHook{cmd/appendix/before}{\def\cref@section@alias{appendix}\def\cref@subsection@alias{appendix}}
\makeatother

\begin{document}
\maketitle

\begin{abstract}

\textit{Mixture-of-Experts} (MoE) models are widely used to scale language models, yet their expert routing behavior and adaptation in a multilingual setting remain underexplored. 
In this work, we study multilingual routing dynamics during continual pre-training of an English-centric MoE model on a multilingual corpus, analyzing how expert usage varies across languages. 
We find that continual multilingual pre-training leads to diffused, language-agnostic routing in early and middle layers, with language specialization primarily emerging in the final layers. 
We also show that token-level vocabulary overlap between languages plays an important role in how languages are routed. 
Motivated by these findings, we propose a parameter-efficient adaptation strategy that updates language-specific and shared experts in the final MoE layers. 
Experiments on MultiBLiMP and Belebele show that our method achieves a strong performance–efficiency trade-off, attaining competitive performance relative to fine-tuning complete final layers, while updating less than $2\%$ of the parameters. 
Overall, our findings provide insights into where and how language specialization emerges in MoEs during continual pre-training and provide practical insights for low-resource multilingual adaptation. 
Our code is available \href{https://github.com/aditi184/moe-routing-adaptation/}{here}.
\end{abstract}
\section{Introduction}
\label{sec:introduction}

\begin{figure}[t]
    \centering
    \includegraphics[width=0.49\textwidth]{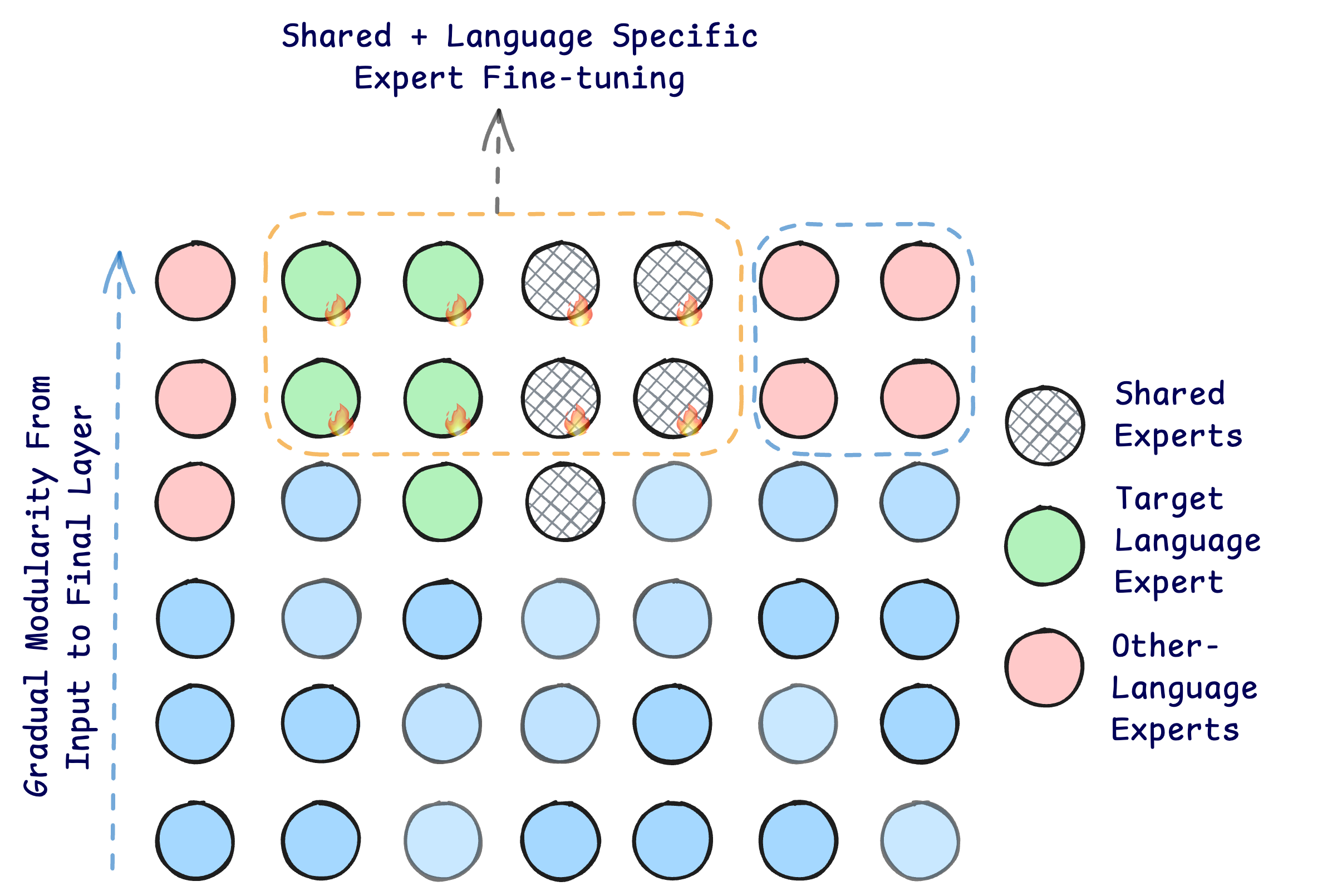}
    \caption{We hypothesize that continually pre-training English-centric MoEs on a mixture of high-resource languages leads to the emergence of language experts. We leverage this specialization for parameter-efficient adaptation to related low-resource languages. }
    \label{fig:teaser}
    \vspace{-0.3cm}
\end{figure}

\textit{Mixture-of-Experts} (MoE) architectures are now the standard for scaling \textit{Large Language Models} (LLMs) because they allow for massive parameter counts while maintaining manageable inference costs \cite{deepseekai2025deepseekv3technicalreport,comanici2025gemini25pushingfrontier,kimiteam2025kimik2openagentic}.
By sparsely activating a subset of the parameters per token, MoE models offer both computational efficiency and a modular structure that makes internal routing behavior amenable to analysis \cite{xue2024openmoeearlyeffortopen,lo-etal-2025-closer,li-etal-2025-decoding}. 
While this modularity has been shown to facilitate specialization in domains such as math or coding \cite{li2022branchtrainmergeembarrassinglyparalleltraining,jiang2024mixtralexperts,muennighoff2024olmoeopenmixtureofexpertslanguage}, the extent to which MoEs develop specialized routing strategies for different languages remains underexplored. At the same time, adapting large MoEs to new languages is computationally demanding. Although their modularity should, in principle, enable {parameter-efficient adaptation} by updating only a subset of experts, it remains unclear which experts to update without first understanding multilingual routing behavior.\looseness-1 

Prior research has explored the internal mechanisms of dense multilingual models.
Findings suggest that they often rely on English as an internal `pivot' or `concept space' in the middle layers \cite{alabi-etal-2024-hidden,wendler2024llamas}, with language-specific neurons primarily localized in the first and last few layers \cite{kojima2024multilingual}.
In contrast, the routing dynamics of multilingual MoE models are less well understood. 
Recent work has begun to address this gap, observing that MoEs exhibit language-agnostic routing in intermediate layers and specialization in early and late layers \cite{bandarkar2025multilingual}. 
However, these analyses are derived predominantly from models trained on English-dominant corpora \cite{abdin2024phi,muennighoff2024olmoeopenmixtureofexpertslanguage,agarwal2025gpt}.

We ask: \textit{does continual pre-training of an MoE model on balanced multilingual data lead to language-specific specialization, 
and if so, can this behavior be leveraged to efficiently adapt a model to related low-resource languages?}

We conduct a controlled study of the routing dynamics of the OLMoE model \cite{muennighoff2024olmoeopenmixtureofexpertslanguage} during multilingual continual pre-training. 
We adapt OLMoE-Base on a 35B-token corpus spanning seven languages. 
Our analysis reveals that routing becomes diffuse and language-agnostic in early and middle layers, with distinct language specialization emerging only in the final layers. 

Leveraging these insights, we investigate whether such language specialization can be used practically for adaptation to low-resource languages. We introduce \textit{Selective and Shared Expert Finetuning} (\texttt{SSFT}), which updates only the relevant language-specific experts and a small set of shared experts in the final layers.
Our results demonstrate that selectively finetuning both specialized and shared experts provides the best trade-off between parameter efficiency and performance. 
This hybrid strategy consistently outperforms the adaptation of specialized experts alone and achieves strong performance while updating a substantially smaller fraction of parameters ($<2\%$ of the model). 

Our core contributions are:
1) \textbf{Multilingual Continual Pre-training Analysis.} With a controlled study of routing dynamics under balanced multilingual training, we show diffused and language-agnostic routing in early to middle layers, with language specialization in final layers. We identify \emph{vocabulary overlap} between languages as an important factor influencing routing behavior.

2) \textbf{Selective and Shared Expert Finetuning (\texttt{SSFT}).} We propose \texttt{SSFT}, showing that both language-specific and shared experts are important for low-resource adaptation. This strategy offers a favorable performance–efficiency trade-off.

The rest of this paper is structured as follows: \S\ref{sec:background} provides background on MoE architectures and the metrics we use to analyze routing dynamics.
\S\ref{sec:routing_analysis} describes our balanced multilingual continual pre-training setup and examines the evolution of routing dynamics. \S\ref{sec:adaptation} proposes and analyzes parameter-efficient adaptation methods. We end by discussing related work in \S\ref{sec:related-work} and conclude in \S\ref{sec:conclusion}.

\section{Preliminaries}
\label{sec:background}

Below, we provide the necessary background on the MoE architecture and the OLMoE model we use, followed by the mathematical framework employed to analyze the routing dynamics. 

\subsection{Mixture-of-Experts}

MoE models extend transformer LMs by replacing the standard feed-forward block with a collection of expert feed-forward networks \cite{shazeer2017outrageouslylargeneuralnetworks,lepikhin2020gshardscalinggiantmodels,fedus2022switchtransformersscalingtrillion}. 
At each MoE layer, a trainable router (a small linear projection followed by a softmax) assigns routing probabilities across all experts. 
For each input token, only the top-$k$ experts with the highest routing probabilities are selected to process it. 
The outputs of the chosen experts are then weighted by their routing probabilities and summed, producing the token representation passed to the next layer.

 We base our experiments on OLMoE-Base\footnote{We use OLMoE due to its fully open-source implementation and publicly available training code, which enables reproducible analysis of routing and expert specialization.}, a decoder-only transformer with MoE feed-forward layers. 
 Each MoE layer consists of 64 experts and a learned router that selects the top 8 experts per token in each layer. 
 Experts are identical in architecture but maintain separate parameters, enabling modularity through routing. 
 OLMoE-Base \cite{muennighoff2024olmoeopenmixtureofexpertslanguage} has been pretrained on approximately 5T predominantly English tokens using a data mixture of the DCLM-Baseline corpus \cite{NEURIPS2024_19e4ea30} and Dolma 1.7 \cite{soldaini2024dolma}. 

\subsection{Analyzing routing behavior}

To study multilingual processing in MoE models, we analyze expert routing patterns across languages and layers to characterize how routing mass is distributed. 
The methods introduced here form the basis for the routing analyses presented in \S\ref{sec:routing_analysis}.

We collect routing information on held-out documents for each language across all decoder layers. For a given \lang{language} $\lang{\ell}$ and \layer{layer} $\layer{k}$, we record post-softmax routing probabilities for all tokens.

Let $E$ denote the number of experts in each MoE layer. 
For the $\doc{i}$-th \doc{document} from language $\lang{\ell}$, let $\tok{T}_\doc{i}$ denote its number of \tok{tokens}. 
We denote by $\mathbf{p}^{(\layer{k})}_{\doc{i},\tok{t}}(\lang{\ell}) \in \Delta^{E-1}$ the routing probability distribution for the $\tok{t}$-th \tok{token} of this document at layer $\layer{k}$.\footnote{This vector contains one element per expert, representing the probability of routing this token to that expert.}
For each document, we compute a document-level expert usage distribution by averaging routing probabilities across tokens:
\vspace{-0.5em}
\[
\mathbf{q}^{(\layer{k})}_\doc{i}(\lang{\ell})
= \frac{1}{\tok{T}_\doc{i}} \sum_{\tok{t}=1}^{\tok{T}_\doc{i}} \mathbf{p}^{(\layer{k})}_{\doc{i},\tok{t}}(\lang{\ell})~. \vspace{-0.5em}
\] 
\noindent This quantity represents the average fraction of routing mass assigned to each expert for a single document, summarizing how that document is routed through the MoE layer. 

We then aggregate document-level distributions to obtain a language-level expert usage distribution for each layer. Let $\doc{N}_\lang{\ell}$ denote the number of documents for language $\lang{\ell}$; the aggregated distribution is given by $\mathbf{q}^{(\layer{k})}(\lang{\ell}) = \frac{1}{\doc{N}_\lang{\ell}} \sum_{\doc{i}=1}^{\doc{N}_\lang{\ell}} \mathbf{q}^{(\layer{k})}_\doc{i}(\lang{\ell})$. This distribution captures the typical expert usage pattern for language $\lang{\ell}$ at layer $\layer{k}$, averaged across documents.

\paragraph{Router entropy.}

To quantify how concentrated expert routing is for a given language, we compute the router entropy at each layer:
\vspace{-0.5em}
\[
H_\layer{k}(\lang{\ell}) = - \sum_{e=1}^{E} \mathbf{q}^{(\layer{k})}(\lang{\ell}){[e]}\, \log \mathbf{q}^{(\layer{k})}(\lang{\ell}){[e]}~. \vspace{-0.5em}
\]
Here, $\mathbf{q}^{(\layer{k})}(\lang{\ell}){[e]}$ denotes the $e$-th element of the vector $\mathbf{q}^{(\layer{k})}(\lang{\ell})$, representing the average routing probability to expert $e$. Lower entropy indicates that tokens from language $\lang{\ell}$ are routed to a small subset of experts, while higher entropy reflects more diffuse routing across experts.

\paragraph{Cross-lingual routing divergence.}

To compare routing behavior between languages, we compute the Jensen--Shannon divergence between their language-level expert usage distributions. For two languages $\lang{\ell_i}$ and $\lang{\ell_j}$ at layer $\layer{k}$, we define
\vspace{-0.5em}
\[
\mathrm{JSD}_\layer{k}(\lang{\ell_i},\lang{\ell_j})
= \mathrm{JSD}\!\left(\mathbf{q}^{(\layer{k})}(\lang{\ell_i}),\, \mathbf{q}^{(\layer{k})}(\lang{\ell_j})\right)~. \vspace{-0.5em}
\]
Low JSD values indicate similar expert usage patterns between the two languages, while higher values indicate more distinct routing behavior.

\subsection{Language selection}
\label{sec:languages}

\textcolor{black}{We use a high- and a low-resource language group for our analyses.}
\textcolor{black}{The high-resource set contains English (en), Arabic (ar), Czech (cs), Spanish (es), Finnish (fi), Hindi (hi), and Russian (ru).
We use this set in \S\ref{sec:routing_analysis} to analyze how expert routing evolves during continual pre-training 
and how these patterns differ from English-centric models.}

\textcolor{black}{The low-resource set contains Catalan (ca), Estonian (et), Marathi (mr), Slovak (sk), Ukrainian (uk), Dutch (nl) and Urdu (ur).}\footnote{Several of these languages are not traditionally considered low-resource \cite{joshi-etal-2020-state}. We treat them as such to maintain a controlled experimental setup and ensure the availability of reliable evaluation benchmarks.}
\textcolor{black}{We use this set in \S\ref{sec:routing_analysis} to analyze cross-lingual co-routing patterns alongside the high-resource set, and in \S\ref{sec:adaptation} to evaluate different adaptation strategies.}
\textcolor{black}{Each low-resource language is paired with a high-resource anchor from the same family, yielding Catalan--Spanish, Estonian--Finnish, Marathi--Hindi, Slovak--Czech, Ukrainian--Russian, and Urdu--Arabic, paired based on token-level vocabulary overlap, script similarity, and the availability of downstream evaluation benchmarks.}\footnote{See Appendix~\ref{app:tokenvocab} for details on token overlap computation.}\footnote{We exclude English--Dutch as an adaptation pair: their token-vocabulary overlap is low ($\sim$9\%) and English routing is too diffuse to isolate language-specific experts.} \textcolor{black}{The resulting pairs span a wide range of overlap, from Spanish--Catalan at approximately 20\% to Hindi--Marathi at over 90\%.}\looseness-1

\section{Routing Dynamics during Continual Pre-training}
\label{sec:routing_analysis}

In a multilingual setting, MoEs potentially allow for experts specializing in individual languages.
However, their internal language-specific or language-agnostic routing behavior remains understudied.
Here, we analyze MoE expert routing during multilingual continual pre-training to assess the emergence of language-specific experts, before leveraging these insights for model adaptation (\S\ref{sec:adaptation}).

\begin{shaded}
\noindent\textbf{Hypothesis.} When an English-centric MoE model is continually pre-trained on balanced multilingual data, expert routing reorganizes to become increasingly language-sensitive, reducing English-dominated routing patterns and exhibiting greater differentiation across languages.
\end{shaded}
\vspace{-0.4cm}

\subsection{Setup and evaluation}

\paragraph{Data.}

For continual pre-training, we sample documents uniformly across high-resource languages (cf.\ \S\ref{sec:languages}) 
from the CulturaX dataset \cite{nguyen2023culturaxcleanedenormousmultilingual}, resulting in a 35B-token training corpus. For our routing analysis, we use 500 held-out validation documents per language (both high- and low-resource) to compute the aggregated language-level expert usage distribution (cf.\ \S\ref{sec:background}).

\begin{figure}[t] 
    \centering 
    \includegraphics[width=0.43\textwidth]{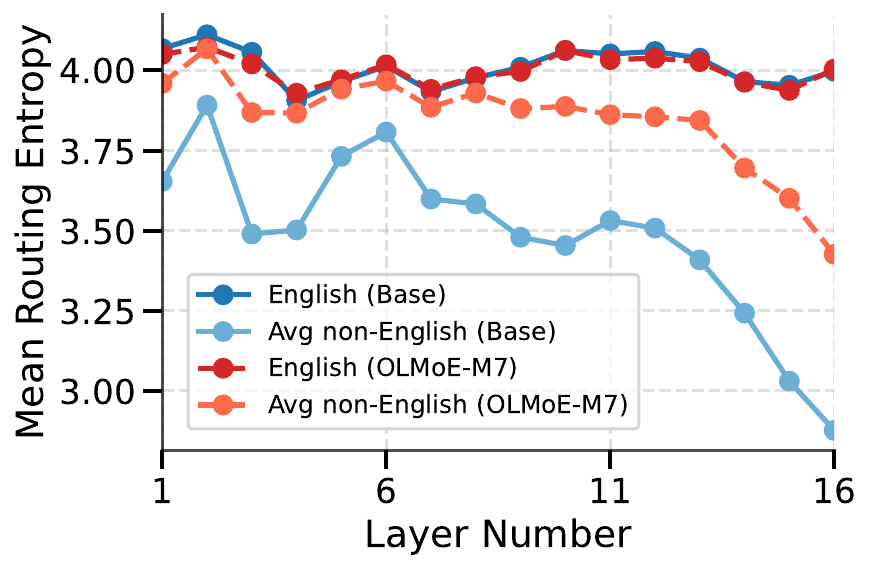} 
    \caption{Comparison of routing entropy across layers for the English and non-English tokens for OLMoE-Base and OLMoE-M7. } 
    \label{fig:entropychange} 
    \vspace{-0.2cm}
\end{figure}

\paragraph{Training setup.}

We perform continual pre-training starting from OLMoE-Base on our 35B-token multilingual corpus and refer to the resulting model as OLMoE-M7. Key architectural details and pre-training hyperparameters for this experiment are summarized in Appendix~\ref{ap:hyperparameters}.  

\paragraph{Evaluation.}

We conduct both intrinsic and extrinsic evaluations to ensure the continual pre-training is successful. 
For intrinsic evaluation, we measure perplexity as a standard indicator of language modeling quality.
For extrinsic evaluation, we evaluate all models on two multilingual downstream benchmarks: \textbf{Belebele} \cite{bandarkar-etal-2024-belebele} and \textbf{MultiBLiMP} \cite{jumelet2025multiblimp10massivelymultilingual}. 
We select these datasets because they cover many languages from the CulturaX pre-training corpus.
Moreover, both are widely used benchmarks to study multilingual model capabilities without requiring task-specific finetuning or instruction tuning, making them well suited for our setup \cite[e.g.,][]{foroutan2025revisitingmultilingualdatamixtures,messmer2025enhancingmultilingualllmpretraining,huang2024survey}. More details can be found in \Cref{app:eval_benchmarks}. 

\paragraph{Results.}

Table~\ref{tab:hrl_all_metrics} in Appendix~\ref{app:hrl_results} presents the intrinsic and extrinsic evaluations. 
Both perplexity and MultiBLiMP performance consistently and substantially improve with continual pre-training; only English degrades slightly, which is expected as the majority of tokens during continual pre-training are non-English. 
On Belebele, which is arguably a more difficult task, we observe much smaller, though still mostly consistent, improvements.

\subsection{Qualitative findings for routing dynamics}

\begin{figure}[t]
    \centering 
    \includegraphics[width=0.43\textwidth]{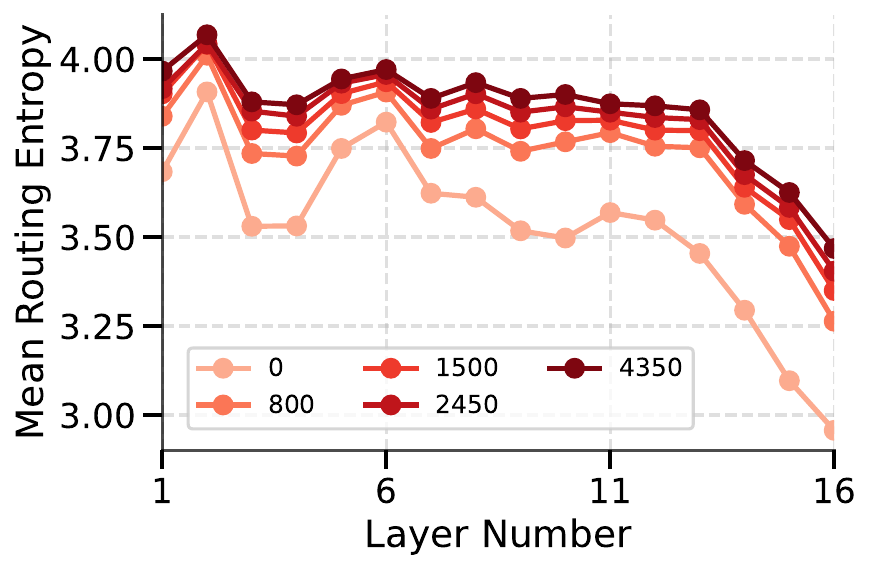} 
    \caption{Routing entropy across layers at different steps of continual pre-training of OLMoE-M7 as indicated in the Legend. Lighter means earlier in step.} 
    \label{fig:entropychangeoverstages} 
\end{figure}

We now analyze the effect of continual pre-training on the model's routing behavior using the metrics introduced in \S\ref{sec:background}: entropy and JSD.
The routing entropy signals how diffused expert activation is per language, but does not directly compare languages one-to-one.
The JSD compares language pairs\, quantifying the extent to which their expert usage differs. 
We use these metrics to compare OLMoE-M7 to the OLMoE-Base model.

\begin{figure*}[t]
    \centering
    \begin{subfigure}[t]{0.36\textwidth}
        \centering
        \includegraphics[width=\textwidth,trim=0pt 0pt 0pt 37pt,clip]{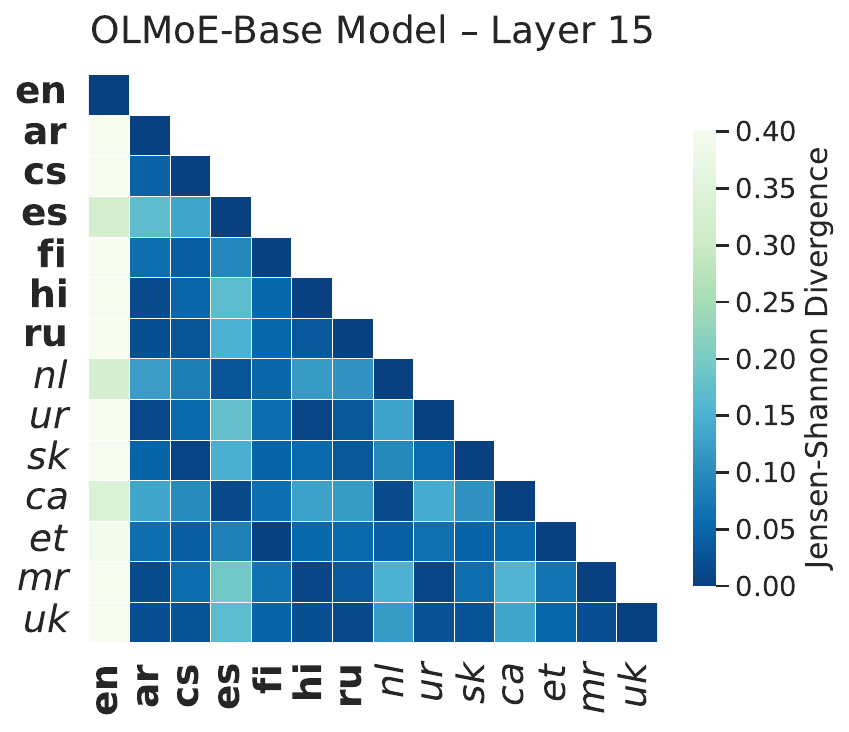}
        \caption{OLMoE-Base - Layer 15}
        \label{fig:jsd_base_l15}
    \end{subfigure}
    \qquad
    \begin{subfigure}[t]{0.36\textwidth}
        \centering
        \includegraphics[width=\textwidth,trim=0pt 0pt 0pt 37pt,clip]{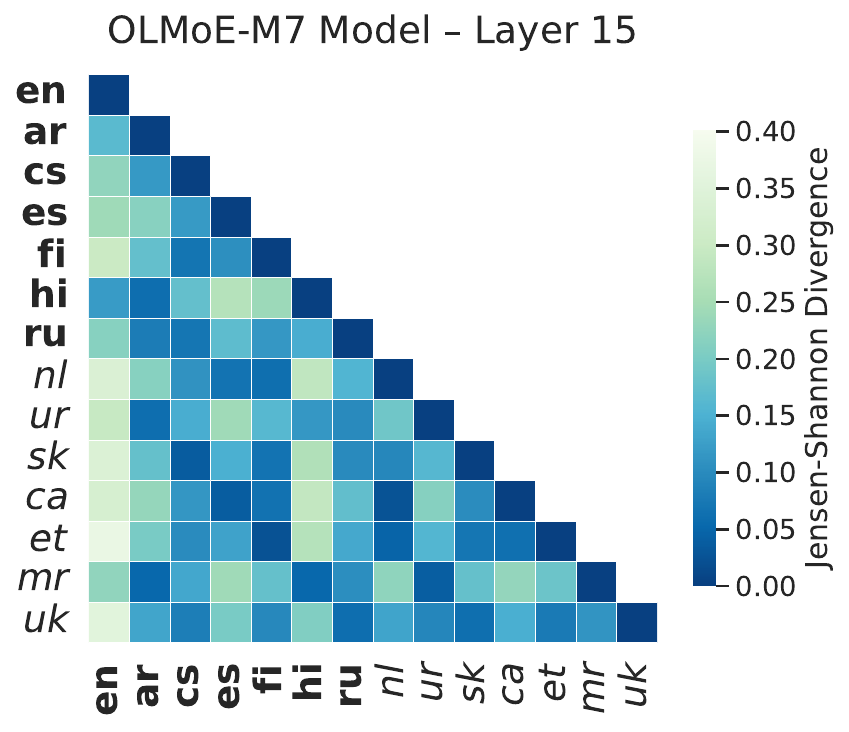}
        \caption{OLMoE-M7 Model - Layer 15}
        \label{fig:jsd_l15}
    \end{subfigure}

    \caption{\textbf{Cross-lingual Routing Divergence in the Final Layer} using Pairwise JSD for OLMoE-Base (left) and OLMoE-M7 (right). Darker blue indicates higher expert sharing. \textbf{Bolded languages} are high-resource; \textit{italicized languages} are low-resource.
    }
    \vspace{-0.3cm}
    \label{fig:jsd_l15_combined}
\end{figure*}


\paragraph{OLMoE has a dedicated routing pattern for English.}

To understand how routing evolves, we first analyze the starting state of the English-centric OLMoE-Base. Looking at entropy in \Cref{fig:entropychange}, we observe that the base model exhibits markedly lower entropy for non-English languages
, particularly in the middle and later layers, which is consistent with findings from \citet{bandarkar2025multilingual}. 
This suggests that in the base model, a relatively small number of experts activates somewhat consistently for non-English languages, particularly in higher layers. This observation is corroborated by the pairwise JSD in the base model, presented in the \Cref{fig:jsd_l15_combined}(a), which confirms that English tokens have very different routing patterns from all non-English languages in the final layer. 
In contrast, the divergence among non-English languages is consistently low, indicating that they are routed through a largely shared set of experts with little distinction between them. This pattern persists across other layers (see \Cref{fig:jsd_l14_l13_base,fig:jsd_l14_l13_stage1} in \Cref{app:routinganalysis}).\looseness-1

\paragraph{Continual pre-training diffuses expert usage across languages and language-specific routing occurs predominantly in final layers.}

When tracking entropy over continual pre-training steps (\Cref{fig:entropychangeoverstages}), we find that the entropy for non-English languages gets closer to (but does not quite approach) the English-level entropy.\footnote{A summary of entropy changes for individual high-resource languages is shown in Figure~\ref{fig:entropy_all_langs} in Appendix~\ref{app:routinganalysis}.} 
Additionally, \Cref{fig:jsdchangeoversteps} shows 
how the JSD changes during training, per layer. 
The between-language divergence is lowest in the first decoder layer; it increases in a non-monotonic manner across the middle layers, and becomes consistently higher towards the final layers. 
Over the course of training, JSD remains largely stable across the middle layers, decreases slightly in the earliest layer, and increases mainly in the final layers. 
Together, these statistics suggest that in the final layers, experts specialize more as evidenced by a larger change in JSD and lower entropy compared to other layers.

Based on these results, we conclude that, contrary to our hypothesis, multilingual continual pre-training does not necessarily induce \textit{language separation} in the expert activation (except for the final layers), although it does \textit{diffuse} them.
Potentially, this is due to the fact that experts activate for multiple (related) languages at a time.

\begin{figure}[t]\centering
    \includegraphics[width=0.41\textwidth]{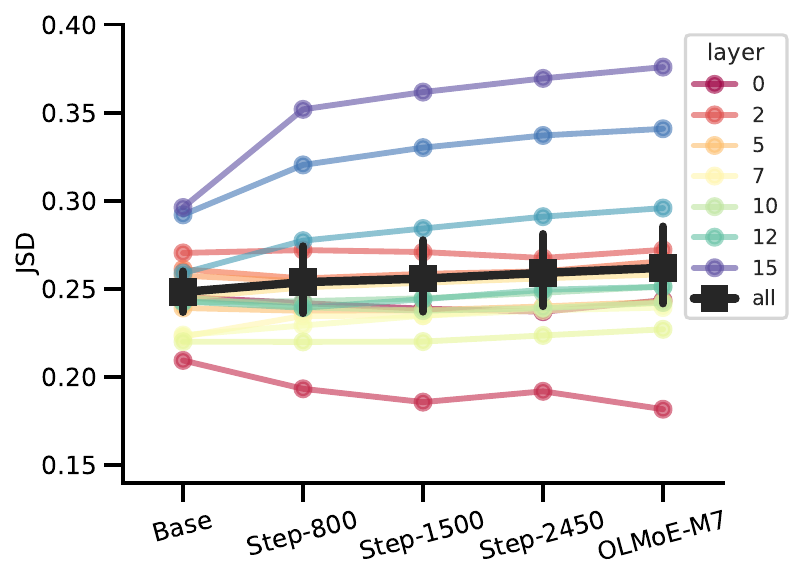}
    \caption{Mean JSD over training steps, showing increased language specialization in final decoder layers.}
    \vspace{-0.3cm}
    \label{fig:jsdchangeoversteps}
\end{figure}

\begin{figure}[t]\centering
    \includegraphics[width=0.43\textwidth]{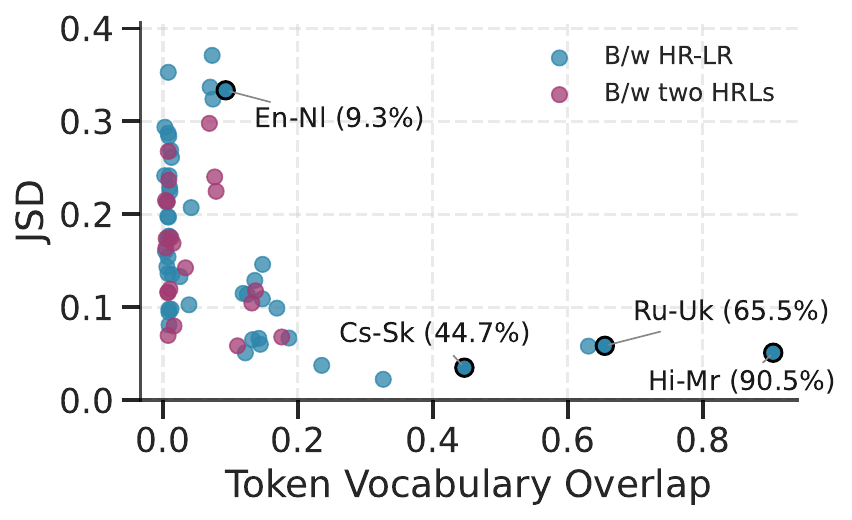}
    \caption{JSD vs Token-Vocabulary Overlap between language pairs. Each point is a language pair and the outlined points with black edges indicate a few qualitative examples for high and low-resource language pairs.}
    \vspace{-0.3cm}
    \label{fig:jsd_vocaboverlap}
\end{figure}

\paragraph{Vocabulary overlap drives differentiation more than language family.}

To better understand differentiation in the final layers, \Cref{fig:jsd_l15_combined}(b) shows a heatmap of pairwise JSD in layer 15. 
Some language pairs exhibit highly similar routing patterns (e.g., Czech–Slovak, Finnish–Estonian), while others are strongly separated (e.g., Hindi–Catalan, or English, which differs from many languages).
\textcolor{black}{In these layers, routing similarity aligns closely with token-level vocabulary overlap, which can sometimes supersede typological markers in driving the router's statistical behavior (e.g., the high routing similarity between Hindi and Marathi vs.\ the divergence between English and Dutch)}.
\textcolor{black}{These pairs share roughly 90\% vs.\ 9\% of tokens in vocabulary, respectively, the latter despite a shared family.} \Cref{fig:jsd_vocaboverlap} plots pairwise JSD against token-level vocabulary overlap for all language pairs.
We find a moderate rank correlation (Spearman’s $\rho{=}{-}0.56$), indicating that higher token overlap corresponds to more similar routing behavior.\footnote{See \Cref{fig:vocaboverlap} in \Cref{app:tokenvocab} for full statistics.} 
Other measures of language similarity---syntactic, phonological, family-based, and geographic similarity from the \texttt{lang2vec} database \citep{littell2017uriel,malaviya17emnlp}---all under-perform in comparison ($\rho$ of $0.30$, $0.36$, $0.11$ and $0.46$, respectively). 
Potentially, statistical vocabulary overlap supersedes the relevance of more traditional markers of language similarity in MoE routing. 
This could mean that prior MoE adaptation methods for low-resource languages
may have been suboptimal \cite{zheng2024efficiently}.\looseness-1 

Finally, we note that \Cref{fig:jsd_l15_combined} demonstrates that English stands out in terms of routing behavior in the base \textit{and} the continually trained model. 
The continual pre-training did reduce English-dominated routing patterns, yet only partially.

\vspace{0.2cm}
\noindent In sum, these analyses indicate that multilingual continual pre-training gradually reshapes routing behavior. 
Rather than inducing clear language-specific expert separation throughout the model, training primarily diffuses expert usage across languages, with more pronounced differentiation in the final layers. In these layers, routing similarity is better explained by vocabulary overlap than by typological or family-based language similarity.

\section{Low-Resource Adaptation}
\label{sec:adaptation}

Motivated by our findings in \S\ref{sec:routing_analysis}, which indicate consistent co-routing of high-overlap language pairs in the final decoder layers,
we now investigate whether the routing dynamics of OLMoE-M7 can be leveraged to adapt the model to related low-resource languages in a parameter-efficient manner.

\begin{shaded}
\noindent\textbf{Hypothesis:} \textcolor{black}{MoE models that are specialized for a high-resource language in the final decoder layers can be leveraged to efficiently improve performance on a related low-resource language.}
\end{shaded}
\vspace{-0.5cm}

\subsection{Adaptation methods}

We first introduce our adaptation methods.

\begin{figure}[t] 
    \centering 
    \includegraphics[width=0.5\textwidth]{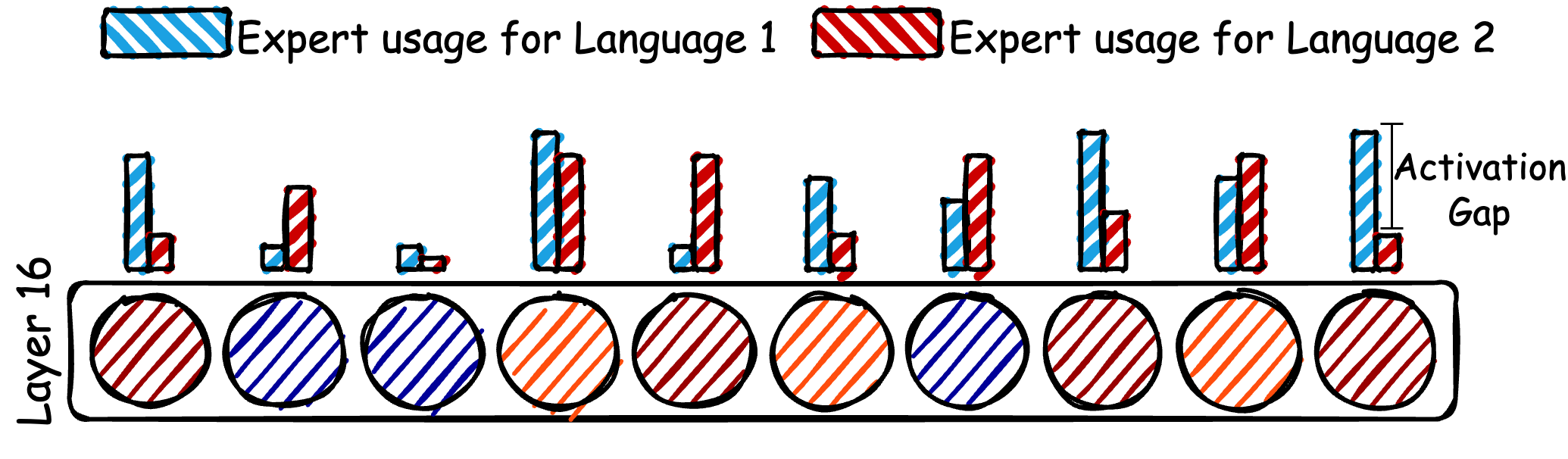} 
    \caption{Illustration of the ``activation gap'' procedure. Specialized experts are identified via the difference in activation frequency between their top two languages.} 
    \vspace{-0.3cm}
    \label{fig:activationgap} 
\end{figure}

\paragraph{Routing-aware expert selection.} 

Testing our hypothesis requires an expert selection procedure, which, given a target language, returns a subset of experts to train.
We propose a method based on what we term \emph{activation gap} (see \Cref{fig:activationgap} for a schematic illustration). 
For each expert and language the model has been continually trained on, we first compute normalized activation frequencies: the fraction of tokens of that language for which the expert is selected among the top-$k$ routed experts at its layer, normalized by the total number of tokens of that language.
Next, we identify the two most activating languages for each expert and compute the activation gap as the difference between their activation frequencies. 
We consider top-2 languages because the second-highest captures the strongest competing language for an expert. 

\paragraph{Selective expert finetuning.}  

We introduce \textit{Selective Expert Finetuning} (\texttt{SEFT}), which adapts only a small subset of experts that are most strongly associated with a target low-resource language.
\texttt{SEFT} adopts the activation gap approach described above to decide which experts are most relevant for the target language. 
Specifically, we select all experts in the final two layers, where our analysis in \S\ref{sec:routing_analysis} shows the strongest language-specific routing differentiation, as evidenced by the highest cross-lingual JSD (\Cref{fig:jsdchangeoversteps})---that exceed an activation gap threshold of 1\% \textcolor{black}{(see Table~\ref{tab:alpha_ablation} in \Cref{app:ablations} for a sensitivity analysis on this threshold)}.
This typically results in selecting 5 to 7 experts per layer.

\paragraph{Selective and shared expert finetuning.} 

For this method, we augment the language-dominant experts selected by \texttt{SEFT} with a small set of experts that are highly active across all languages and term this approach \textit{Selective and Shared Expert Finetuning} (\texttt{SSFT}).
Specifically, for each of the final two layers we add five such shared experts for each layer to the finetuning pool and update them jointly with the language-specific experts. \textcolor{black}{We ablate the number of shared experts $k$ in \Cref{app:ablations}, Table~\ref{tab:k_ablation}}.

\paragraph{Baselines.} 

To validate that gains from \texttt{SEFT} and \texttt{SSFT} arise from meaningful expert specialization rather than parameter count alone, we introduce a \textit{Random Expert Finetuning} (\texttt{Random-SEFT}) baseline, where a randomly selected set of experts matched in number to \texttt{SEFT} is finetuned. 
To ensure that improvements are not simply due to finetuning a larger number of experts, we conduct an additional control experiment called \texttt{SEFT-Top20}. 
Here, we expand the \texttt{SEFT} expert set using the same activation-gap ranking to include approximately 30\% of experts (top-20) in each of the final two layers. 
This allows us to disentangle the effects of expert quantity from expert specificity.
As another baseline, we also consider \textit{All-Experts Finetuning} (\texttt{AEFT}), where all experts and router parameters in the final two layers are updated, testing whether broader expert adaptation provides additional benefits beyond selective specialization. 
As an upper bound, we include \textit{Full-Model Finetuning} (\texttt{Full-FT}), in which all model parameters are updated.
\\ \\
\noindent Crucially, across all parameter-efficient strategies, only the selected experts and routers for respective layers are trainable. All other experts, attention layers, and embeddings remain frozen. These methods update approximately 75 to 251M parameters. In contrast, \texttt{AEFT} updates 800M parameters, while \texttt{Full-FT} updates all 7B parameters. 

\subsection{Experimental Setup}

Next, we detail the experimental setup for our low-resource adaptation experiments.

\paragraph{Data.}

\textcolor{black}{For each low-resource target we assume access to a related high-resource ``anchor'' language seen during continual pre-training, and use the anchor's routing behavior to identify which experts to adapt. We adopt the six low-resource targets and their high-resource anchors introduced in \S\ref{sec:languages}.} To simulate a low-resource setting, we sample approx.\ 300M tokens per language, corresponding to roughly 5\% of the multilingual pretraining budget.

\paragraph{Hyperparameters.}

We 
sweep the learning rate over the set $\{1e{-}5,1e{-}4,4e{-}4,1e{-}3,4e{-}3\}$ for all methods. 
Model selection is based on perplexity measured on a held-out validation set for each low-resource language. 
For all methods, we report results for the checkpoint achieving the lowest validation perplexity.

\subsection{Results}

We report results on the two multilingual benchmarks introduced in \S\ref{sec:routing_analysis}, MultiBLiMP and Belebele (cf.\ \Cref{app:eval_benchmarks} for details).

\paragraph{Finetuning language-specific experts outperforms finetuning random experts.}  

\begin{figure}[t] 
    \centering 
    \includegraphics[width=0.48\textwidth,trim=0 0 0 0]{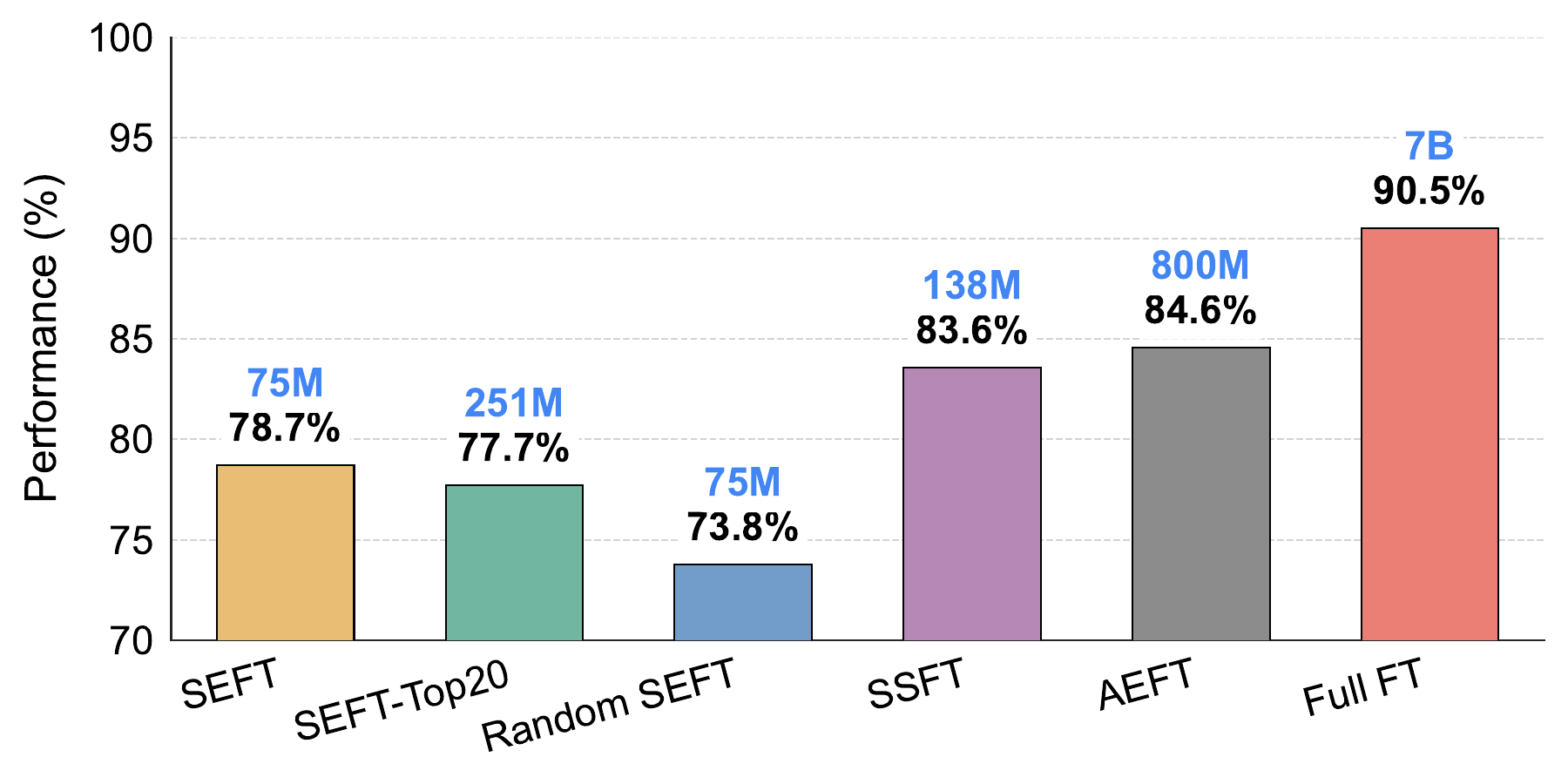} 
    \caption{Average \textbf{MultiBLiMP} performance across target languages comparing different adaptation strategies (\texttt{SEFT}, \texttt{SSFT}) against baselines. Numbers in blue next to each bar indicate trainable parameters.} 
    \label{fig:baselinecomparison} 
\end{figure}

We first validate the effectiveness of our routing-aware expert selection finetuning by comparing \texttt{SEFT} to \texttt{Random-SEFT}, which adapt the same number of parameters.
On MultiBLiMP, \Cref{fig:baselinecomparison} shows that \texttt{Random-SEFT} yields the lowest average performance across all target languages ($73.8\%$), substantially lagging behind \texttt{SEFT} ($78.7\%$) and other routing-aware strategies.

\paragraph{Shared experts improve overall performance.}

Targeting language-specific experts provides gains over \texttt{Random-SEFT}, and the remaining results in \Cref{fig:baselinecomparison} furthermore show that incorporating shared experts further improves performance:
\texttt{SSFT} yields an average performance of $83.6\%$, outperforming \texttt{SEFT} by $4.9$ percentage points, \textcolor{black}{with gains scaling monotonically in the number of shared experts $k$ (Table~\ref{tab:k_ablation} in Appendix~\ref{app:ablations})}.
Looking at individual languages, in \Cref{tab:multiblimp_results}, \texttt{SSFT} leads to substantial performance improvements for, e.g., Marathi ($+7.2$ points) and Estonian ($+6.9$). 
To test whether performance gains are driven simply by the increased numbers of trainable parameters or by the nature of the experts, we compare \texttt{SSFT} against \texttt{SEFT-top20}, a baseline where we increase the number of language-specific experts updated without including shared experts. 
Despite updating a larger budget of experts, \texttt{SEFT-Top20} achieves only $77.7\%$ average accuracy, falling well short of \texttt{SSFT} ($83.6\%$). 
On Belebele (\Cref{tab:belebeleresults}), \texttt{SSFT} again outperforms \texttt{SEFT} on average, with the largest gains for Estonian ($+4.2$ points) and Catalan ($+1.9$).

\paragraph{Computational advantage.}

For OLMoE-1B, full \texttt{Full-FT} on 300M tokens requires approximately 2 hours on 16 H100 GPUs ($\approx$ 32 GPU-hours), corresponding to $\approx 9.8\times10^{17}$ FLOPs. 
In contrast the parameter-efficient adaptation methods, \texttt{SSFT} and \texttt{SEFT} update only 75-138M parameters and complete training in 45-50 minutes on 4 H100s ($\approx$ 3.3 GPU hours), utilizing $\approx 9.8\times10^{15}$ FLOPs. 
This leads to a 10x reduction in GPU-hours and 100x reduction in FLOPs highlighting \texttt{SSFT}'s computational efficiency in comparison to \texttt{Full-FT}.\footnote{While Full-FT provides the overall strongest performance when adapting on 300M tokens, we note that it leads to catastrophic forgetting for the languages adapted to in the initial continual pre-training stage when training on > 800M tokens in the subsequent stage of adaptation. \texttt{SEFT} on the other hand, leads to less catastrophic forgetting. We provide further details on these results in \Cref{app:catastrophicforgetting}.}

\begin{table}[t]
\centering
\small
\begin{tabular}{lcccc|l}
\toprule
\textbf{Lang.} & \textbf{SEFT} & \textbf{SSFT} & \textbf{AEFT} & \textbf{Full FT} & \textbf{M7} \\
\midrule
ca  & 82.9 & 85.4 & 89.0 & 93.3 & 81.3 \\
et  & 68.4 & 75.3 & 74.0 & 81.9 & 61.5 \\
mr  & 66.5 & 73.7 & 74.6 & 81.7 & 68.5 \\
sk  & 86.3 & 88.6 & 89.9 & 92.7 & 86.0 \\
uk  & 81.7 & 85.1 & 85.5 & 91.9 & 80.7 \\
ur  & 86.5 & 93.3 & 94.4 & 96.5 & 83.3 \\
\bottomrule
\end{tabular}
\caption{Results on \textbf{MultiBLiMP} for target languages using different adaptation strategies.}
\label{tab:multiblimp_results}
\end{table}

\section{Related Work}
\label{sec:related-work}

\paragraph{Multilingual MoE Analysis.} 

Prior work has studied expert specialization in multilingual MoE models, though evidence for language-specific modularity remains mixed and is largely limited to encoder-based or sequence-to-sequence architectures. 
\citet{zoph2022st} find that encoder MoEs tend to specialize over shallow token groupings rather than by language, casting doubt on whether multilingual MoEs can reliably form language-specific experts. 
In multilingual neural MT, \citet{kudugunta2021exploring} propose routing strategies at multiple granularities, but do not analyze expert specialization across languages. 
Similarly, Meta’s NLLB MoE models demonstrate the scalability of MoEs for translation, yet do not investigate how experts specialize or how routing patterns vary across languages \cite{nllbteam2022languageleftbehindscaling}. 
\citet{zheng2024efficiently} leveraged MoE modularity primarily for medical domain specialization, introducing language-family experts and hybrid routing strategies that utilize late-stage language specialization but rely on explicit language identity at inference time. 
\citet{bandarkar2025multilingual} study routing dynamics in MoE-based LLMs, showing that experts are language-specific in early and late layers but largely shared in middle layers.
They analyze pretrained models only and propose a steering method which is applied during inference to improve multilingual generalization. In contrast, we focus on training-based adaptation.
\textcolor{black}{Complementary to these 
analyses, \citet{kallini-etal-2025-false} study how subword vocabulary overlap shapes cross-lingual representation alignment and transfer in bilingual models; we instead probe the mechanical routing decisions that overlap induces in multilingual MoEs.}

\paragraph{Efficient Multilingual Adaptation.}

Previous work aims to mitigate the `curse of multilinguality' through modularity and resource optimization. 
Frameworks like MAD-X \cite{pfeiffer-etal-2020-mad} and MAFT \cite{alabi-etal-2022-adapting} utilize adapters and continual pre-training to adapt models to new languages, while \citet{ansell2022composable} perform adaptation via parameter-efficient sparse masks. 
\citet{marchisio-etal-2023-mini} propose `mini-model' training to reduce compute costs, and \citet{gurgurov-etal-2024-adapting} integrate external knowledge graphs to compensate for low resource languages.
\begin{table}[t]
    \centering
    \small
    \begin{tabular}{lcccc|cc}
        \toprule
        \textbf{Lang.} & \textbf{SEFT} & \textbf{SSFT} & \textbf{AEFT} & \textbf{Full FT} & \textbf{M7} \\
        \midrule
        ca     & 29.2 & 31.1 & 29.2 & 34.9 & 31.4 \\
        et    & 26.0 & 30.2 & 29.6 & 30.6 & 25.3 \\
        mr     & 29.9 & 27.7 & 26.2 & 30.9 & 25.8 \\
        sk      & 33.8 & 34.3 & 33.7 & 34.6 & 33.0 \\
        uk  & 31.6 & 31.3 & 32.2 & 35.0 & 30.3 \\
        ur        & 28.3 & 28.9 & 29.2 & 27.8 & 28.9 \\
        \bottomrule
    \end{tabular}
    \caption{\textbf{Belebele} (4-shot) performance for target languages using different adaptation strategies.}
    \label{tab:belebeleresults}
\end{table}

\section{Conclusion}
\label{sec:conclusion}
\vspace{-0.1cm}
We studied the routing dynamics of MoE models in multilingual continual pre-training. 
Our analysis reveals that multilingual adaptation leads to diffuse, language-agnostic routing throughout the early and middle layers of the network.
Distinct language specialization emerges gradually in the final layers, where co-routing of languages correlates more strongly with token-level vocabulary overlap than with language families.
\textcolor{black}{While our experiments focus on the OLMoE architecture, the routing dynamics and specialization trends across layers align with findings from MoE and dense model literature \citep{bandarkar2025multilingual,kojima2024multilingual}.}\looseness-1

Leveraging these insights, we proposed Selective and Shared Expert Finetuning (\texttt{SSFT}), a parameter-efficient adaptation strategy.
By updating only the language-dominant and shared experts in the final layers, \texttt{SSFT} achieves strong performance on benchmarks like MultiBLiMP and Belebele, while updating less than $2\%$ of the model parameters. 
Hence \texttt{SSFT} offers computational advantages compared to full fine-tuning and can be more robust to catastrophic forgetting. 
Overall, our findings suggest that effective low-resource adaptation of MoE models relies on both targeting specialized experts for new languages while also preserving shared experts to maintain cross-lingual stability.
\textcolor{black}{While this work focuses on low-resource adaptation, the observed dynamics of specialization and cross-lingual expert sharing are broadly applicable to the study of multilingual MoEs and could naturally inform more general cross-lingual transfer strategies or efficient modular model adaptation.}
\section*{Limitations}
\label{sec:limitations}

Due to the substantial computational demands of continual pre-training, we concentrated our experiments on OLMoE-Base architecture (1B active / 7B total parameters). Validating these findings across varying model scales remains an avenue for future work.
Our proposed adaptation strategy (\texttt{SSFT}) relies on the existence of a high-resource ``anchor'' language with significant vocabulary overlap to identify relevant experts. This approach may prove less effective for language isolates or low-resource languages that lack a close high-resource relative in the pre-training data. 
We examine routing dynamics specifically within the context of continual pre-training on a balanced multilingual corpus (35B tokens). The emergence of language specialization in the final layers may differ under alternative training regimens, such as curriculum learning, different data mixing ratios, or during pre-training from scratch.

\section*{Acknowledgements}
We thank the members of our McGill and Mila research labs for their feedback throughout the course of this project. We especially thank Jay Gala and Harman Singh for their thoughtful reviews and valuable discussions that helped improve this work. Finally, we thank the Compute Canada and Mila IT support teams for their continuous assistance and for providing the computational resources necessary to run our experiments.

\bibliography{custom}
\appendix
\clearpage

\section{Pre-training Hyperparameters}
\label{ap:hyperparameters}
\Cref{tab:hyperparms} provides the detailed hyperparameters and architectural specifications used for the continual pre-training of the OLMoE-Base model.

\begin{table}[h]
\centering
\small
\setlength{\tabcolsep}{3.5pt} 
\begin{tabular}{ll}
\toprule
\textbf{Hyperparameter} & \textbf{Value} \\
\midrule
Model architecture & Decoder-only MoE \\
Number of layers & 16 \\
Hidden dimension ($d_{\text{model}}$) & 2048 \\
Attention heads & 16 \\
Activation function & SwiGLU \cite{shazeer2020gluvariantsimprovetransformer} \\
Sequence length & 4096 \\
Vocabulary size & 50,280 \\
\midrule
Number of experts per layer & 64 \\
Top-$k$ routing & 8 \\
MoE routing type & Dropless, sparse MLP \\
Router z-loss weight & 0.001 \cite{zoph2022st} \\
Auxiliary MoE loss weight & 0.01 \cite{shazeer2017outrageouslylargeneuralnetworks} \\
\midrule
Optimizer & AdamW \\
Learning rate & $1 \times 10^{-4}$ \\
Adam $\beta$ coefficients & (0.9, 0.95) \\
Weight decay & 0.1 \\
\midrule
Global batch size & 2048 \\
Precision & bfloat16 (AMP) \\
\bottomrule
\end{tabular}
\caption{Key hyperparameters used for continually pretraining OLMoE-Base.}
\label{tab:hyperparms}
\end{table}

\begin{table*}[t]
\centering
\small
\begin{tabular}{lccccccccc}
\toprule
\textbf{Language} & \textbf{English} & \textbf{Arabic} & \textbf{Czech} & \textbf{Spanish} & \textbf{Finnish} & \textbf{Hindi} & \textbf{Russian} & \textbf{Target} & \textbf{Original Target} \\
\midrule
OLMoE-M7 & 97.7 & 96.0 & 94.0 & 97.1 & 96.3 & 99.1 & 96.0 & - & - \\
\midrule
\multicolumn{10}{c}{\textit{SSFT Results}} \\
\midrule
Catalan   & 98.2 & 95.3 & 93.9 & 95.1 & 95.6 & 99.2 & 95.8 & 85.9 & 81.3 \\
Estonian  & 98.1 & 94.9 & 93.1 & 95.6 & 94.5 & 99.1 & 95.2 & 77.0 & 61.5 \\
Marathi   & 96.4 & \textbf{80.7} & \textbf{74.4} & \textbf{88.0} & \textbf{81.4} & \textbf{97.2} & \textbf{71.1} & 79.1 & 68.5 \\
Slovak    & 97.9 & 95.1 & 93.0 & 96.3 & 96.0 & 99.2 & 95.6 & 89.0 & 86.0 \\
Ukrainian & 97.9 & 95.1 & 92.6 & 97.0 & 95.8 & 99.2 & 95.0 & 83.5 & 80.7 \\
Urdu      & 97.9 & 94.2 & 93.3 & 97.2 & 96.3 & 99.0 & 95.6 & 92.4 & 83.3 \\
\midrule
\multicolumn{10}{c}{\textit{Full FT Results}} \\
\midrule
Catalan   & 96.9 & 94.8 & \textbf{91.6} & \textbf{94.3} & \textbf{93.1} & 99.4 & 96.2 & 95.8 & 81.3 \\
Estonian  & 97.7 & 95.1 & \textbf{90.9} & \textbf{94.6} & \textbf{86.1} & 98.9 & 95.7 & 90.9 & 61.5 \\
Marathi   & 97.4 & 94.0 & 93.3 & 96.6 & \textbf{94.6} & \textbf{91.7} & 95.7 & 89.1 & 68.5 \\
Slovak    & 97.5 & 94.2 & \textbf{90.5} & \textbf{96.0} & 93.7 & 99.0 & 96.4 & 93.8 & 86.0 \\
Ukrainian & 97.4 & 95.1 & 92.8 & 96.8 & 95.4 & 99.3 & 92.1 & 93.0 & 80.7 \\
Urdu      & 97.7 & \textbf{89.5} & 94.3 & 96.8 & 96.2 & 99.2 & 95.6 & 96.7 & 83.3 \\
\bottomrule
\end{tabular}
\caption{MultiBlimp performance on High-Resource languages used to train OLMoE-M7 and effect of SSFT and Full Finetuning by training on 800M tokens. Highlighted cells denote cases where performance drops by more than one standard deviation of the downstream performance relative to OLMoE-M7 continually trained on high-resource languages.}
\label{tab:catastrophic_forgetting}
\end{table*}

\section{Continual Pre-training Results}
\label{app:hrl_results}
\textcolor{black}{Table~\ref{tab:hrl_all_metrics} reports per-language intrinsic (perplexity) and extrinsic (MultiBLiMP, Belebele 4-shot) evaluations for OLMoE-Base and OLMoE-M7 across the seven high-resource pre-training languages. Continual multilingual pre-training substantially reduces perplexity and improves MultiBLiMP performance on all non-English languages, with a modest degradation on English; Belebele improvements are smaller but mostly consistent.}

\begin{table*}[h]
\centering
\small
\begin{tabular}{lcccccc}
\toprule
\textbf{Language} &
\multicolumn{2}{c}{\textbf{Perplexity $\downarrow$}} &
\multicolumn{2}{c}{\textbf{MultiBLiMP $\uparrow$}} &
\multicolumn{2}{c}{\textbf{BeleBele (4-shot) $\uparrow$}} \\
\cmidrule(lr){2-3} \cmidrule(lr){4-5} \cmidrule(lr){6-7}
 & \textbf{OLMoE-Base} & \textbf{OLMoE-M7} & \textbf{OLMoE-Base} & \textbf{OLMoE-M7} & \textbf{OLMoE-Base} & \textbf{OLMoE-M7} \\
\midrule
English (eng)  & \textbf{12.48} & 13.36 & \textbf{0.981} & 0.977 & \textbf{0.590} & 0.390 \\
Arabic (arb)   & 5.68  & \textbf{3.00}  & 0.828 & \textbf{0.960} & 0.307 & \textbf{0.344} \\
Czech (ces)    & 11.33 & \textbf{4.45}  & 0.758 & \textbf{0.940} & \textbf{0.354} & \textbf{0.354} \\
Spanish (spa)  & 10.18 & \textbf{6.78}  & 0.928 & \textbf{0.971} & 0.309 & \textbf{0.364} \\
Finnish (fin)  & 17.82 & \textbf{5.01}  & 0.753 & \textbf{0.963} & 0.316 & \textbf{0.328} \\
Hindi (hin)    & 3.20  & \textbf{1.97}  & 0.930 & \textbf{0.991} & 0.296 & \textbf{0.328} \\
Russian (rus)  & 7.04  & \textbf{4.22}  & 0.870 & \textbf{0.960} & 0.354 & \textbf{0.368} \\

\bottomrule
\end{tabular}
\caption{Performance on high-resource languages before and after multilingual continual pretraining. OLMoE-M7 improves performance across all non-English languages, with a modest degradation for English.}
\label{tab:hrl_all_metrics}
\end{table*}

\section{Token-Vocabulary Overlap}
\label{app:tokenvocab}
\begin{figure}[t]
    \centering
    \includegraphics[width=0.5\textwidth]{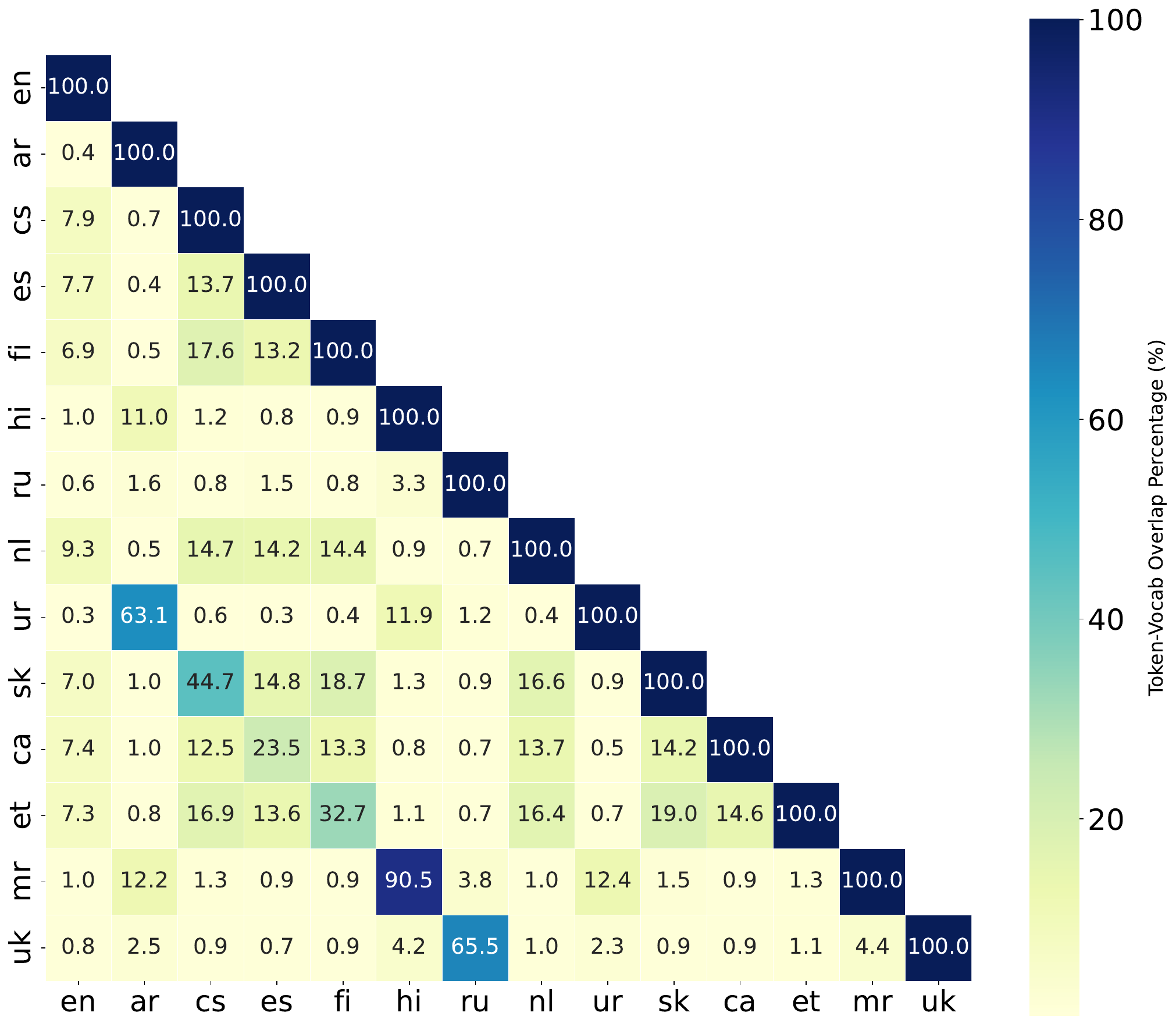}
    \caption{Token-Vocabulary Overlap across language pairs.}
    \label{fig:vocaboverlap}
\end{figure}

We compute token-level vocabulary overlap by tokenizing parallel Bible chapters using the OLMoE tokenizer, leveraging the wide multilingual coverage of the Bible. Pairwise vocabulary overlap statistics are shown in Figure~\ref{fig:vocaboverlap}.

\section{Routing Analysis}
\label{app:routinganalysis}
\begin{figure*}[t]
    \centering
    \begin{subfigure}[t]{0.49\textwidth}
        \centering
        \includegraphics[width=\textwidth]{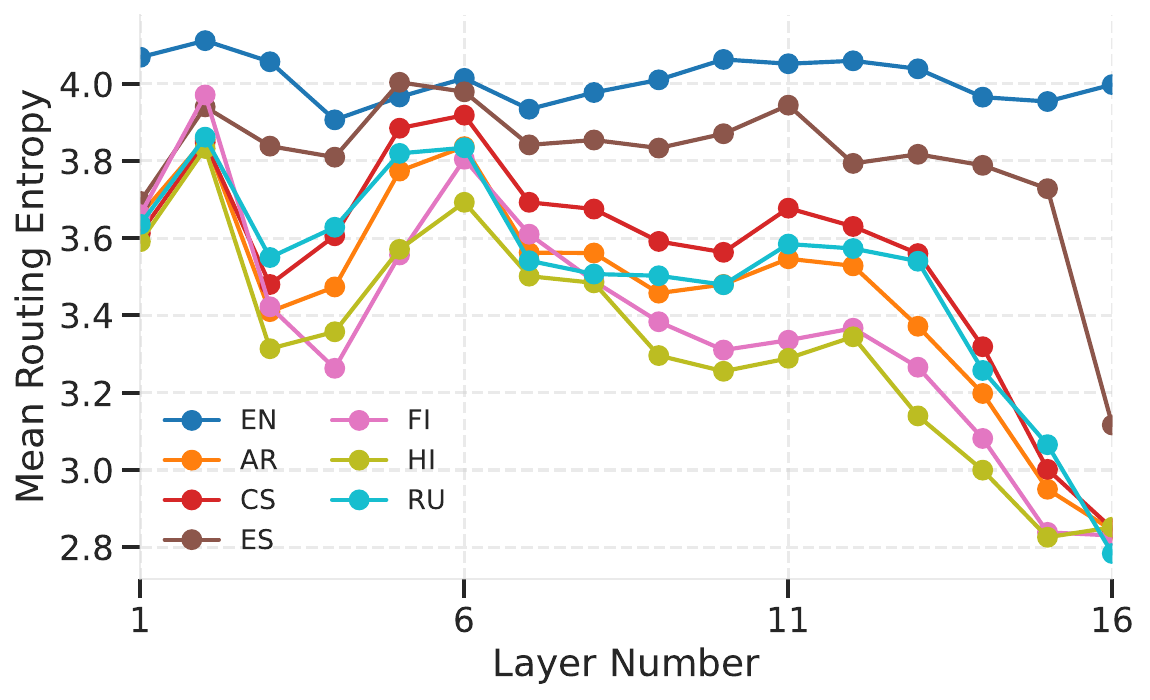}
        \caption{OLMoE-Base}
        \label{fig:entropyall_base}
    \end{subfigure}
    \hfill
    \begin{subfigure}[t]{0.49\textwidth}
        \centering
        \includegraphics[width=\textwidth]{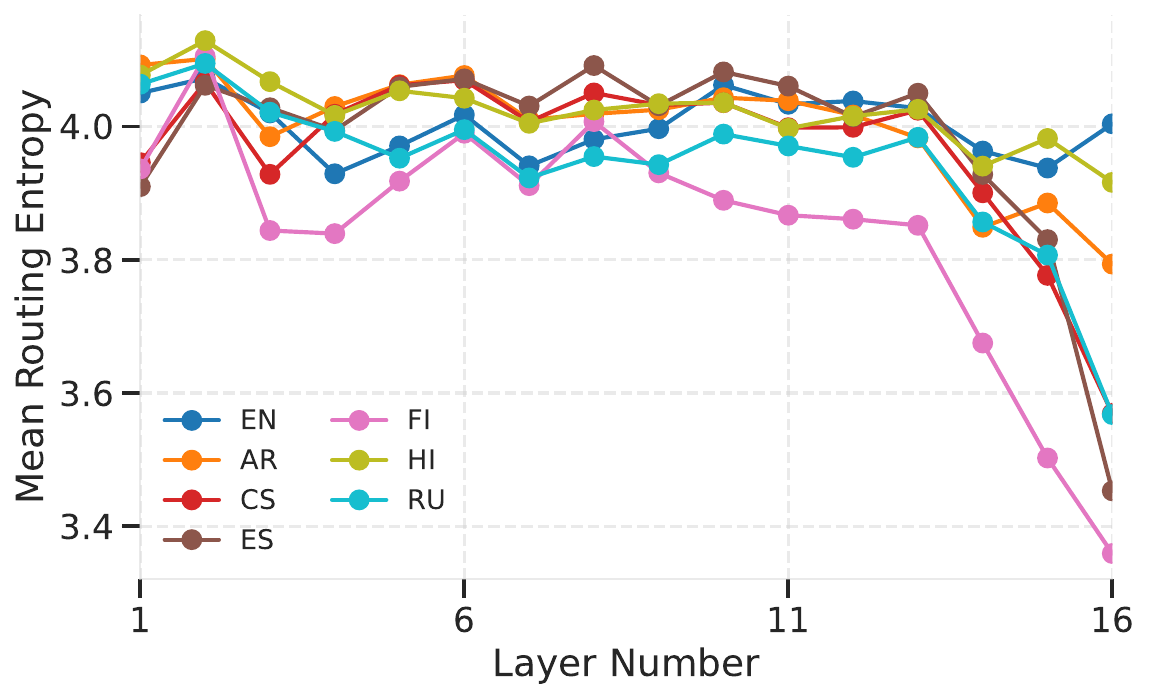}
        \caption{OLMoE-M7}
        \label{fig:entropyall_stage1}
    \end{subfigure}

    \caption{Comparison of routing entropy across layers for OLMoE-Base (left) and OLMoE-M7 (right) across all high-resource languages.}
    \label{fig:entropy_all_langs}
\end{figure*}

\begin{figure*}[t]
    \centering
    \begin{subfigure}[t]{0.4\textwidth}
        \centering
        \includegraphics[width=\textwidth,trim=0pt 0pt 0pt 37pt,clip]{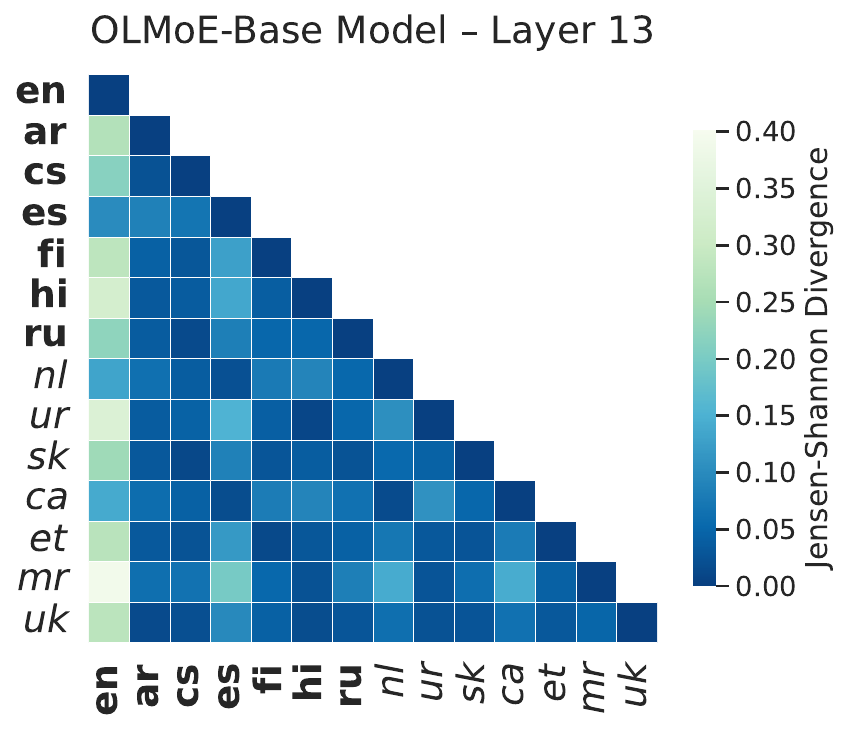}
        \caption{OLMoE-Base - Layer 13}
        \label{fig:jsd_base_l13}
    \end{subfigure}
    \begin{subfigure}[t]{0.4\textwidth}
        \centering
        \includegraphics[width=\textwidth,trim=0pt 0pt 0pt 37pt,clip]{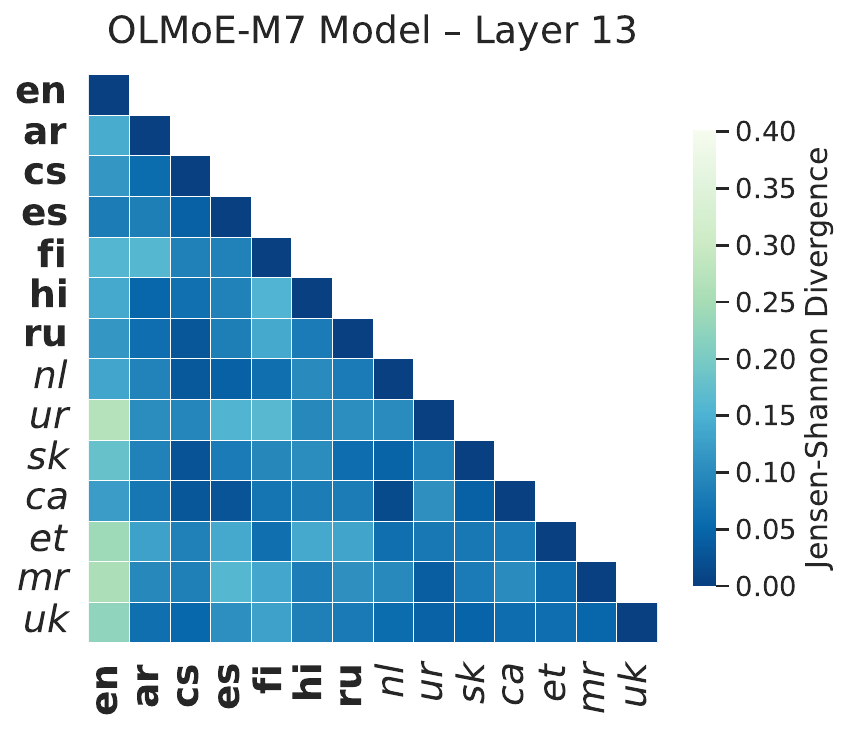}
        \caption{OLMoE-M7 - Layer 13}
        \label{fig:jsd_m7_l13}
    \end{subfigure}

    \caption{\textbf{Cross-lingual Routing Divergence in the Layer 13} using Pairwise Jensen-Shannon Divergence (JSD)for OLMoE-Base (left) and OLMoE-M7 (right). Darker blue indicates higher expert sharing.}
    \label{fig:jsd_l14_l13_base}
\end{figure*}

\begin{figure*}[t]
    \centering
    \begin{subfigure}[t]{0.4\textwidth}
        \centering
        \includegraphics[width=\textwidth,trim=0pt 0pt 0pt 37pt,clip]{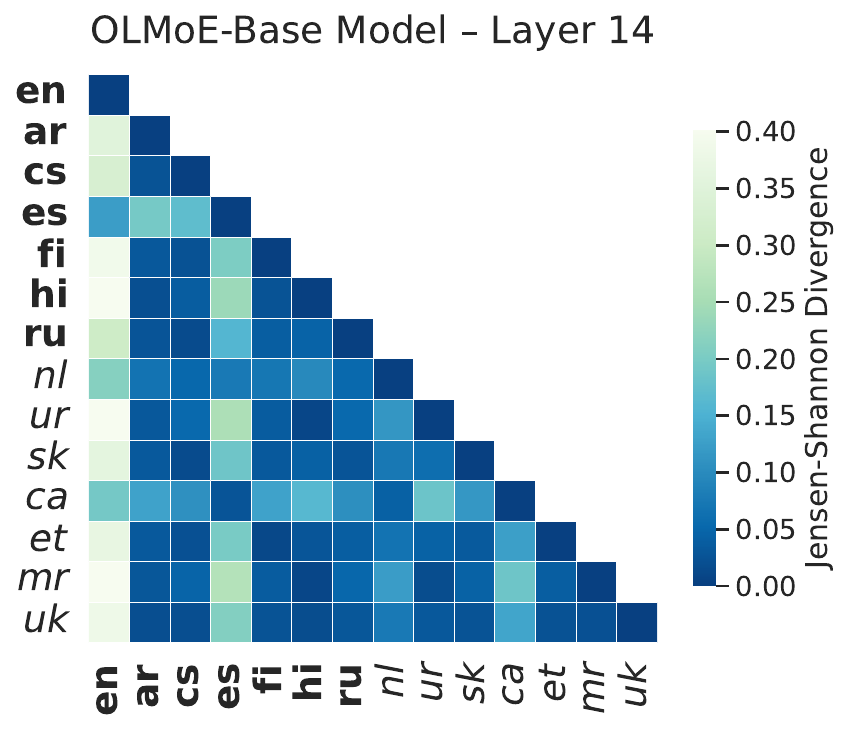}
        \caption{OLMoE-Base - Layer 14}
        \label{fig:jsd_base_l14}
    \end{subfigure}
    \begin{subfigure}[t]{0.4\textwidth}
        \centering
        \includegraphics[width=\textwidth,trim=0pt 0pt 0pt 37pt,clip]{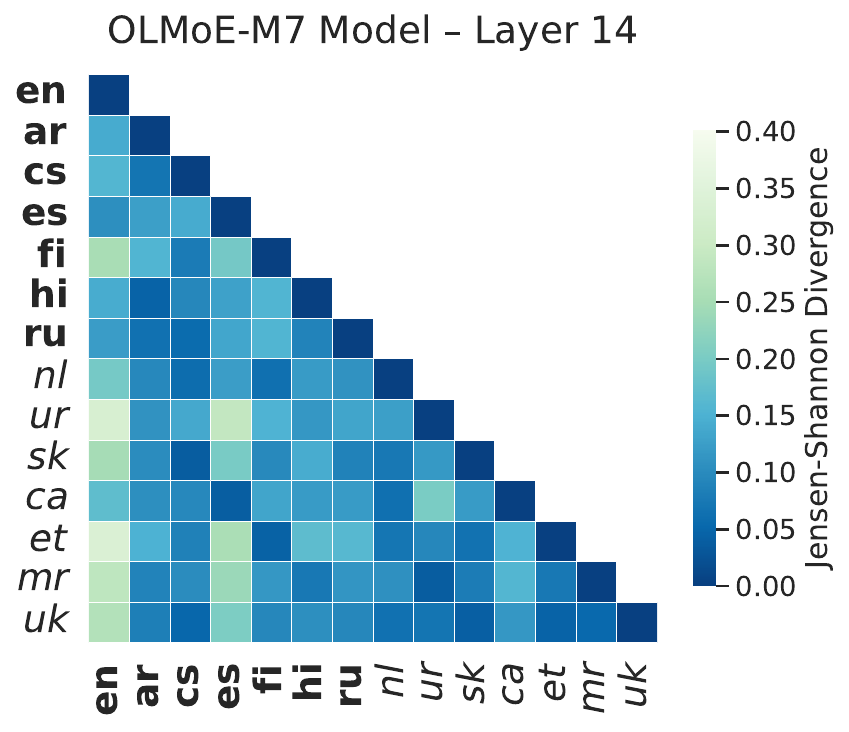}
        \caption{OLMoE-M7 - Layer 14}
        \label{fig:jsd_m7_l14}
    \end{subfigure}

    \caption{\textbf{Cross-lingual Routing Divergence in the Layer 14} using Pairwise Jensen-Shannon Divergence (JSD) for OLMoE-Base (left) and OLMoE-M7 (right). Darker blue indicates higher expert sharing.}
    \label{fig:jsd_l14_l13_stage1}
\end{figure*}

We provide additional routing analysis specifically for the OLMoE-Base model. Figures~\ref{fig:jsd_l14_l13_base} and ~\ref{fig:jsd_l14_l13_stage1} illustrate the pairwise Jensen-Shannon Divergence (JSD) for the base model and OLMoE-M7 in layers 13 and 14, supplementing the analysis in \Cref{sec:routing_analysis}. Additionally, \Cref{fig:entropy_all_langs} illustrates the routing entropy across different layers for OLMoE-Base and OLMoE-M7 models across high-resource languages.

\section{Evaluation Benchmarks}
\label{app:eval_benchmarks}
We evaluated our adaptation strategies on the Belebele and Multiblimp:
Belebele is a multiple-choice machine reading comprehension benchmark covering 122 language variants, with questions grounded in short passages from FLORES-200 \cite{nllbteam2022languageleftbehindscaling}, enabling direct cross-lingual comparison of semantic understanding. 
MultiBLiMP is a multilingual benchmark of syntactic and morphological minimal pairs generated using Universal Dependencies and UniMorph, evaluating sensitivity to fine-grained grammatical distinctions across languages.

\section{Sensitivity to Expert Selection Hyperparameters}
\label{app:ablations}

\textcolor{black}{We provide two ablations supporting the hyperparameter choices in \texttt{SEFT} and \texttt{SSFT}: the activation gap threshold $\alpha$ (\Cref{tab:alpha_ablation}) and the number of shared experts $k$ (\Cref{tab:k_ablation}).}

\textcolor{black}{\paragraph{Activation gap threshold ($\alpha$).}
The 1\% activation gap acts as a filter: if it is too low, we admit ``noisy'' experts that lack language-specific dominance (similar to our \texttt{SEFT-Top20} control); if it is too high, the selected expert pool shrinks dramatically and few experts remain per language, yielding little change relative to the baseline. \Cref{tab:alpha_ablation} reports \texttt{SEFT} performance on MultiBLiMP when varying $\alpha \in \{2\%, 1\%, 0.0001\%\}$. The strict 2\% setting leaves the model with too few experts, most evidently for Marathi, where no experts meet the threshold, while the relaxed 0.0001\% setting fails to construct a meaningful expert pool, yielding inconsistent gains. The $\alpha = 1\%$ setting used in the main paper provides the best balance.}

\begin{table}[h]
\centering
\small
\begin{tabular}{lcccc}
\toprule
Lang. & $\alpha{=}2\%$ & $\alpha{=}1\%$ & $\alpha{=}0.0001\%$ & M7 \\
\midrule
Cat. & 81.9 & 82.9 & 83.5 & 81.3 \\
Est. & 65.0 & 68.4 & 72.8 & 61.5 \\
Mar. & n/a  & 66.5 & 67.4 & 68.5 \\
Slk. & 85.7 & 86.3 & 86.8 & 86.0 \\
Ukr. & 79.6 & 81.7 & 80.4 & 80.7 \\
Urd. & 82.2 & 86.5 & 82.2 & 83.3 \\
\bottomrule
\end{tabular}
\caption{MultiBLiMP performance for \texttt{SEFT} under different activation gap thresholds $\alpha$. ``M7'' is the OLMoE-M7 baseline before adaptation. ``n/a'' indicates that no experts met the threshold (Marathi at $\alpha{=}2\%$).}
\label{tab:alpha_ablation}
\end{table}

\textcolor{black}{\paragraph{Number of shared experts ($k$).}
The shared experts in \texttt{SSFT} are identified by computing the mean activation of every expert across all seven high-resource pre-training languages, using a held-out validation set of 5{,}000 samples per language; we then select the $k$ experts with the highest mean activation. To assess sensitivity, we vary $k \in \{0, 1, 3, 5\}$ (with $k{=}0$ recovering \texttt{SEFT}). \Cref{tab:k_ablation} shows a clear monotonic improvement as $k$ grows, confirming that shared experts contribute cross-lingual transfer beyond what language-specific experts provide alone. Combined with the \texttt{SEFT-Top20} control in \Cref{fig:baselinecomparison}, this indicates that the \emph{sharedness} of these experts, not the added parameter count, drives the improvement.}

\begin{table}[h]
\centering
\small
\begin{tabular}{lccccc}
\toprule
Lang. & $k{=}0$ & $k{=}1$ & $k{=}3$ & $k{=}5$ & M7 \\
\midrule
Cat. & 82.9 & 83.2 & 84.2 & 85.4 & 81.3 \\
Est. & 68.4 & 71.5 & 72.9 & 75.3 & 61.5 \\
Mar. & 66.5 & 67.0 & 72.4 & 73.7 & 68.5 \\
Slk. & 86.3 & 86.5 & 88.0 & 88.6 & 86.0 \\
Ukr. & 81.7 & 79.3 & 78.1 & 85.1 & 80.7 \\
Urd. & 86.5 & 84.7 & 86.4 & 93.3 & 83.3 \\
\bottomrule
\end{tabular}
\caption{MultiBLiMP performance for \texttt{SSFT} under varying numbers of shared experts $k$ ($k{=}0$ corresponds to \texttt{SEFT}). ``M7'' is the OLMoE-M7 baseline before adaptation.}
\label{tab:k_ablation}
\end{table}

\section{Catastrophic Forgetting}
\label{app:catastrophicforgetting}

Having established SSFT as an effective and parameter-efficient adaptation strategy, we next examine whether it also improves training stability over full-finetuning. Using an extended 800M-token adaptation setting to amplify forgetting effects, we observe that full-model finetuning substantially degrades performance on MultiBlimp previously seen languages during continual pretraining, whereas SSFT largely preserves performance on those languages as shown in Table~\ref{tab:catastrophic_forgetting}. This suggests that selective experts finetuning acts as a regularization enabling adaptation without overwriting already-learnt representations.

\section{Licenses of Scientific Artifacts}
\label{app:licenses}

\textcolor{black}{We list the licenses of the scientific artifacts used in this work. The CulturaX dataset is released under the CC0-1.0 and ODC-BY licenses. The OLMoE model is released under the Apache 2.0 license. Our use of these artifacts is consistent with their intended use and licensing terms.}

\end{document}